\documentclass[11pt,a4paper]{article}
\usepackage[hyperref]{emnlp2020}

\usepackage{times}
\usepackage{latexsym}
\usepackage{tabularx}
\usepackage{booktabs}
\usepackage{graphicx}
\usepackage{hyperref}
\usepackage{xcolor}
\usepackage{soul}
\usepackage{amsfonts}
\usepackage{stmaryrd}
\usepackage[T1]{fontenc}
\usepackage{enumitem}
\usepackage{hyperref}
\usepackage{pifont}
\usepackage{afterpage}
\usepackage{amsmath}
\usepackage{amssymb}
\usepackage{multirow} 
\usepackage{balance}
\usepackage{subcaption}
\usepackage{stfloats}

\aclfinalcopy

\usepackage{microtype}

\newcommand{\cmark}{\ding{51}}%
\newcommand{\xmark}{\ding{55}}%

\definecolor{darkspringgreen}{rgb}{0.09, 0.45, 0.27}
\definecolor{darkred}{rgb}{0.55, 0.0, 0.0}

\definecolor{alizarin}{rgb}{0.82, 0.1, 0.26}
\definecolor{ao(english)}{rgb}{0.0, 0.5, 0.0}
\definecolor{cadmiumgreen}{rgb}{0.0, 0.42, 0.24}

\newcommand{\complex}{ComplEx}
\newcommand{\distmult}{DistMult}
\newcommand{\transe}{TransE}
\newcommand{\transh}{TransH}

\newcommand{\fb}{\texttt{FB15K-Wiki}} % \scalebox{.9}{+}}}

\newcommand{\wn}{\texttt{WN18RR}}

\newcommand{\kge}{KGE}

\newcommand*\pct{\scalebox{.9}{\%}}

\newcommand{\cwa}{CWA}
\newcommand{\owa}{OWA}

\title{Evaluating the Calibration of Knowledge Graph \\ Embeddings for Trustworthy Link Prediction}
\author{
  Tara Safavi\thanks{\enspace This work was done during an internship at Bloomberg.} \\
  University of Michigan \\
  tsafavi@umich.edu \\ 
  \And
  Danai Koutra \\
  University of Michigan \\ 
  dkoutra@umich.edu \\
  \And
  Edgar Meij \\
  Bloomberg \\
  emeij@bloomberg.net \\
}

\date{}

\begin{document}
\maketitle
\begin{abstract}
Little is known about the trustworthiness of predictions made by knowledge graph embedding (\kge) models. 
In this paper we take initial steps toward this direction by investigating the \emph{calibration} of \kge{} models, or the extent to which they output confidence scores that reflect the expected correctness of predicted knowledge graph triples. 
We first conduct an evaluation under the standard closed-world assumption (\cwa), in which predicted triples not already in the knowledge graph are considered false, and show that existing calibration techniques are effective for \kge{} under this common but narrow assumption. 
Next, we introduce the more realistic but challenging open-world assumption (\owa),
in which unobserved predictions are not considered true or false until ground-truth labels are obtained. 
Here, we show that existing calibration techniques are much less effective under the \owa{} than the \cwa{}, and provide explanations for this discrepancy. 
Finally, to motivate the utility of calibration for \kge{} from a practitioner's perspective, we conduct a unique case study of human-AI collaboration, showing that calibrated predictions can improve human performance in a knowledge graph completion task. 
\end{abstract}

\section{Introduction}
\label{sec:intro}
Knowledge graphs are essential resources in natural language processing tasks such as question answering and reading comprehension~\cite{shen2019multi, yang2019enhancing}. %  and language modeling~\cite{logan2019barack}. 
Because they are by nature incomplete, extensive research efforts have been invested into completing them via different techniques~\cite{ji2020survey,BelthZVK20}. 

\begin{figure}[t!]
    \centering
    \includegraphics[width=0.75\columnwidth]{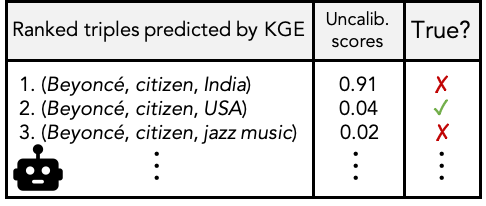}
    \caption{An example of how optimizing for ranking does not necessarily lead to trustworthy prediction scores. 
    Here, an uncalibrated \kge{} model would perform well according to ranking metrics because the true triple is ranked highly, even though it receives a much lower score than the incorrect top-ranked triple and a similar score to the nonsensical triple below it.  
    % This problem is especially prevalent in link prediction under the open-world assumption (\S~\ref{sec:open}). 
    }
    \vspace{-.3cm}
    \label{fig:example}
\end{figure}

One such technique is knowledge graph embedding (\textbf{\kge}), which involves learning latent representations of entities and relations to be used toward predicting new facts. 
\kge{} models are most commonly optimized for the \textbf{link prediction} task, which tests their ability to ``learn to rank'' plausible knowledge graph triples higher than implausible ones. 
While \kge{} accuracy as measured by ranking-based link prediction metrics has been steadily improving on benchmark datasets over the past decade~\cite{ruffinelli2020you}, such evaluation setups can be misleading.
As shown in Figure~\ref{fig:example}, ranking only considers the \emph{ordering} of prediction scores, so models can perform well according to ranking metrics even if they assign high scores to incorrect or nonsensical triples~\cite{wang2019evaluating}. 

As such, the practical utility of \kge{} for real-world knowledge graph completion remains limited, especially given that other completion techniques such as relation extraction and manual curation have already been reliably deployed in commercial and scientific settings~\cite{suchanek2007yago, dong2014knowledge, ammar2018construction}.
The outstanding issue that we address relates to \emph{trustworthiness}:
That is, to what degree can one trust the predictions made by \kge{}? 
We believe that trustworthiness is an important part of making \kge{} practical for knowledge graph completion. 

In this paper we propose to investigate \textbf{confidence calibration} as a technique toward making \kge{} more trustworthy. 
Intuitively, calibration is a post-processing step that adjusts \kge{} link prediction scores to be representative of actual correctness probabilities~\cite{guo2017calibration}. 
Calibration has several benefits. 
From the \emph{systems} perspective, natural language processing pipelines that include knowledge graphs can rely on calibrated confidence scores to determine which \kge{} predictions to trust. 
From a \emph{practitioner}'s perspective, calibrated confidence scores act as decision support for accepting or verifying \kge{} predictions.
Toward this direction we contribute the following: 

\paragraph{Task}
We evaluate \kge{} calibration for link prediction, which is important for making \kge{} viable for deployment. 
While many knowledge graph embedding models exist, their calibration and general trustworthiness are under-explored (\S~\ref{sec:related}). 

\paragraph{Complementary evaluations}
We first evaluate the calibration of established \kge{} models 
under the commonly-used closed-world assumption (\textbf{\cwa{}}), in which triples not present in the knowledge graph are considered false (\S~\ref{sec:closed}). 
We show that existing calibration techniques are highly effective for \kge{} under this assumption. 
Next, we introduce the more challenging open-world assumption (\textbf{\owa{}}), which reflects how \emph{practitioners} would use \kge{}: Triples not present in the knowledge graph are assumed to be \emph{unknown}, rather than false, until ground-truth labels are obtained (\S~\ref{sec:open}). 
We show that existing calibration techniques are less effective under the \owa{} than the \cwa{}, and provide explanations for this discrepancy.

\paragraph{Case study}
Finally, as a proof of concept on the benefits of \kge{} calibration, we conduct a case study in which data annotators complete knowledge graph triples with the help of \kge{} predictions. 
We show that presenting calibrated confidence scores alongside predictions significantly improves human accuracy and efficiency in the task, motivating the utility of calibration for human-AI tasks.

\section{Related work}
\label{sec:related}

While knowledge graph embeddings and calibration have both been extensively studied in separate communities---see~\cite{ji2020survey,ruffinelli2020you} for reviews of \kge{} and~\cite{guo2017calibration} for an overview of calibration for machine learning---relatively little work on calibration \emph{for} knowledge graph embeddings exists.

In the domain of relation extraction, a few works calibrate predicted knowledge graph triples as components of large-scale relation extraction systems. 
\citet{dong2014knowledge} used Platt scaling~\cite{platt1999probabilistic} to calibrate the probabilities of factual triples in the proprietary Knowledge Vault dataset, and \citet{west2014knowledge} used Platt scaling in a search-based fact extraction system. 
However, we focus on link prediction with \kge{} models that learn only from the knowledge graph itself (\S~\ref{sec:kge}). 

We are aware of only two recent works that investigate calibration for \kge{}, both of which address the task of triple classification~\cite{tabacof2019probability,pezeshkpour2020revisiting}.
By contrast, we focus on link prediction, which is a different---and much more common~\cite{safavi2020codex}---\kge{} evaluation task. 
We also contribute an evaluation under the open-world assumption, whereas~\citet{tabacof2019probability} evaluate under the closed-world assumption only.
Finally, unique to our work, we conduct a human-AI case study to demonstrate the benefits of calibration from a practitioner's perspective. 

\section{Preliminaries}
\label{sec:prelim}
\begin{table*}[ht!]
\centering
\caption{Scoring functions of models used in our evaluation. \textbf{Bold letters} indicate vector embeddings. 
$+$ indicates that the scoring function is translational, and
$\times$ indicates that the scoring function is bilinear.
}
\label{table:models}
\resizebox{1\textwidth}{!}{
\begin{tabular}{ r c l l } 
\toprule
      & Type & Scoring function $f$ & Scoring function notes  \\
\toprule
	{\transe{}}~\cite{bordes2013translating} & $+$ & $- \| \mathbf{h} + \mathbf{r} - \mathbf{t}\|$ &  We use the L$_2$ norm \\
	{\transh{}}~\cite{wang2014knowledge} &  $+$ & $- \| \mathbf{h_{\perp}} + \mathbf{r} - \mathbf{t_{\perp}}\|$ & Projects $\mathbf{h}$, $\mathbf{t}$ onto relation-specific hyperplanes to get $\mathbf{h_{\perp}}$, $\mathbf{t_{\perp}}$  \\
	{\distmult{}}~\cite{yang2014embedding} & $\times$ & $\mathbf{h}^{\top} \text{diag}(\mathbf{r}) \mathbf{t}$ & diag($\cdot$) turns a vector into a diagonal matrix  \\ 
	{\complex{}}~\cite{trouillon2016complex} & $\times$ & $\text{Re}\left(\mathbf{h}^{\top} \text{diag}(\mathbf{r}) \mathbf{\overline{t}}\right)$ & $\mathbf{\overline{t}}$: Complex conjugate of $\mathbf{t}$; Re: Real part of a complex number \\
 \bottomrule
\end{tabular}
}
\end{table*}

\subsection{Knowledge graph embeddings}
\label{sec:kge}

A knowledge graph $G$ comprises a set of entities $E$, relations $R$, and (\emph{head}, \emph{relation}, \emph{tail}) triples $(h, r, t) \in E \times R \times E$.
A knowledge graph embedding (\textbf{\kge}) takes triples $(h, r, t)$ as input and learns corresponding embeddings $(\mathbf{h}, \mathbf{r}, \mathbf{t})$ to maximize a scoring function $f : E \times R \times E \rightarrow \mathbb{R}$, such that more plausible triples receive higher scores. 

\paragraph{Models}
In this paper we consider four \kge{} models: 
\textbf{\transe}~\cite{bordes2013translating}, \textbf{\transh}~\cite{wang2014knowledge}, \textbf{\distmult}~\cite{yang2014embedding}, and \textbf{\complex}~\cite{trouillon2016complex}.
Table~\ref{table:models} gives the scoring function of each model. 

We choose these models because they are efficient and representative of two main classes of \kge{} architecture---translational (\transe{}, \transh{}) and bilinear (\distmult{}, \complex{})---which allows us to interpret how different types of scoring functions affect calibration. 
Moreover, these earlier models tend to be used by NLP practitioners.
For example, the language model in~\cite{logan2019barack} uses \transe{} embeddings, and the machine reading system in~\cite{yang2019enhancing} uses \distmult{} embeddings.
From our knowledge of the literature, practitioners using \kge{} are more likely to use earlier, established models.
Since our work targets real-world applications, we prioritize such models. 

\subsection{Link prediction}
\label{sec:lp}

The link prediction task, which is most commonly used to evaluate \kge{}~\cite{safavi2020codex}, is conducted as follows. 
Given a test triple $(h, r, t)$, we hold out one of its entities or its relation to form a query $(h, r, ?)$, $(?, r, t)$, or $(h, ?, t)$.
The model then scores all tail entities $t_i \in E$, head entities $h_i \in E$, or relations $r_i \in R$ as answers to the respective query such that higher-ranked completions $(h, r, t_i)$, $(h_i, r, t)$, or $(h, r_i, t)$ 
are more plausible. 
Prior to computing rankings, all true triples across train, validation, and test beyond the given test triple are filtered out~\cite{bordes2013translating}. 

Under the closed-world assumption (\textbf{\cwa}, \S~\ref{sec:closed}), models are evaluated by their ability to score the true test triples $(h, r, t)$ as high as possible, because it is assumed that all triples not seen in the knowledge graph are incorrect.
Under the open-world assumption (\textbf{\owa}, \S~\ref{sec:open}), models simply score all predicted completions, and the predictions not seen in the knowledge graph are not considered true or false until ground-truth labels are obtained.

\subsection{Confidence calibration} 
\label{sec:calibration}

In the context of link prediction, calibration is the extent to which a \kge{} model outputs probabilistic confidence scores that reflect its expected accuracy in answering queries. 
For example, for 100 predicted triple completions scored at a confidence level of 0.99 by a perfectly calibrated model, we expect 99 of these predictions to be correct.
% according to some binary link prediction metric. 
% We discuss such metrics in \S~\ref{sec:closed-task}.

Calibration is a \emph{post-processing} step. 
To calibrate a \kge{} model, separate calibration parameters are learned on a held-out validation set using the prediction scores of the uncalibrated model. 
These parameters do not affect the trained, fixed embeddings, but rather transform the model's scores. 

\paragraph{Negative samples}
All calibration methods require negatives to appropriately adjust prediction scores for plausible and implausible triples. 
However, link prediction benchmarks (\S~\ref{sec:closed-data}) do not contain negatives.
Therefore, per positive instance, we assume that only the held-out entity or relation correctly answers the query, and take all other completions as negative samples.

This approach, which has been shown to work well in practice for training \kge{} models~\cite{ruffinelli2020you}, treats link prediction as \emph{multiclass}: The ``class'' for each query is its true, held-out entity or relation. 
Since this approach is less faithful to reality for queries that have many entities as correct answers, 
in this paper we evaluate calibration for the \textbf{relation prediction} task---that is, answering $(h, ?, t)$ queries---because there are usually fewer correct answers to relation queries than entity queries.\footnote{For example, in our \fb{} dataset (\S~\ref{sec:closed-data}), the mean number of relations between each unique pair of entities is 1.12, and the median is 1.}
While the methods we describe are general, 
for brevity we focus on relation prediction in this rest of this section. 

\subsection{Calibration techniques}
\label{sec:calibration-techniques}

Let $(h, ?, t)$ be a relation query, $k = |R|$ be the number of relations in the graph, and
\begin{align}
    \mathbf{z} = [f(h, r_{1}, t), \hdots, f(h, r_{k}, t)]^{\top} \in \mathbb{R}^{k}
    \label{eq:uncalib}
\end{align}
be a vector of uncalibrated \kge{} prediction scores across all relations $r_i \in R$, such that $z_i = f(h, r_i, t)$.
Note that for head or tail queries $(?, r, t)$ or $(h, r, ?)$, $\mathbf{z}$ would instead contain prediction scores across all entities in $E$. 

Our goal is to learn a function that transforms the uncalibrated score vector $\mathbf{z}$ into \emph{calibrated probabilities} $\mathbf{z}' \in \mathbb{R}^{k}$.  
Post-calibration, the final predicted answer $\hat{r}$ to the query and corresponding confidence score $\hat{p}$ are taken as
\begin{align}
    \hat{r} = \arg\max{[\mathbf{z}']} \, \textrm{ and } \, \hat{p} = \max{[\mathbf{z}']},
\end{align}
where $\hat{p}$ reflects the expectation that $\hat{r}$ correctly answers the query, i.e., is the ``class'' of the query. 

\paragraph{One-versus-all}
One approach to multiclass calibration is to set up $k$ one-versus-all \emph{binary} calibration problems, and combine the calibrated probabilities for each class afterward. 
The classic \textbf{Platt scaling} technique~\cite{platt1999probabilistic}, which was originally designed for binary classification, can be extended to the multiclass setting in this manner. 
For each class, scalar parameters $a$ and $b$ are learned such that the calibrated probability of the query belonging to the $i$-th class is given by $\hat{p}_i = \sigma_{\textrm{sig}}(az_{i} + b)$, where $\sigma_{\textrm{sig}}$ denotes the logistic sigmoid. 
The parameters are optimized with negative log-likelihood (i.e., binary cross-entropy) loss, which is standard for obtaining probabilistic predictions~\cite{niculescu2005predicting}.
Afterward, all $\hat{p}_i$ are gathered into $\mathbf{z}' = [\hat{p}_1, \hdots, \hat{p}_k]$ and normalized to sum to 1. 

Another well-known calibration technique that fits in the one-versus-all framework is \textbf{isotonic regression}~\cite{zadrozny2002transforming}. 
For each class, a nondecreasing, piecewise constant function $g$ is learned to minimize the sum of squares $[\mathbf{1}(r_i) - g(\sigma_{\textrm{sig}}(z_i))]^2$ across all queries,
where $\mathbf{1}(r_i)$ is 1 if the class of the given query is $r_i$ and 0 otherwise. 
The calibrated probability of the query belonging to the $i$-th class is taken as $\hat{p}_i = g(\sigma_{\textrm{sig}}(z_i))$.
Again, these scores are gathered into $\mathbf{z}' = [\hat{p}_1, \hdots, \hat{p}_k]$ and normalized to sum to 1.

\paragraph{Multiclass}
An alternative approach is to use the softmax $\sigma_{\textrm{sm}}$ to directly obtain probabilities over $k$ classes, rather than normalizing independent logistic sigmoids. 
To this end, \citet{guo2017calibration} propose a variant of Platt scaling that learns weights $\mathbf{A}~\in~\mathbb{R}^{k \times k}$ and biases $\mathbf{b} \in \mathbb{R}^k$ to obtain calibrated confidences $\mathbf{z}' = \sigma_{\textrm{sm}}(\mathbf{A}\mathbf{z} + \mathbf{b})$. 
$\mathbf{A}$ and $\mathbf{b}$ are optimized with cross-entropy loss. 

The weight matrix $\mathbf{A}$ can either be learned with the full $k^2$ parameters (\textbf{matrix scaling}), or can be restricted to be diagonal (\textbf{vector scaling}).
We compare both approaches in \S~\ref{sec:closed}.

\section{Closed-world evaluation}
\label{sec:closed}
We first evaluate \kge{} calibration under the \textbf{closed-world assumption} (\cwa), in which we assume triples not observed in a given knowledge graph are false. 
This assumption, which is standard in \kge{} evaluation~\cite{ruffinelli2020you}, helps narrow evaluation down to a well-defined task in which models are judged solely by their ability to fit known data. 
It is therefore important to first explore this (restrictive) assumption before moving to the more realistic but challenging \owa{} (\S~\ref{sec:open}). 

\subsection{Task and metrics}
\label{sec:closed-task}

As described in \S~\ref{sec:lp}, link prediction under the \cwa{} is conducted by constructing queries from test triples and evaluating models' abilities to score these test triples as high as possible.
We measure accuracy by the proportion of top-ranked predicted relations that correctly answer each query.\footnote{Here we use top-1 accuracy because there are relatively few relations in knowledge graphs. However, any binary link prediction metric (i.e., hits@$k$) may be used.}

We quantify a \kge{} model's level of calibration with expected calibration error (\textbf{ECE})~\cite{guo2017calibration}.
ECE measures the degree to which a model's confidence scores match its link prediction accuracy in bins partitioning [0, 1]. 
Given $M$ such bins of equal size, ECE is defined as $\sum_{m=1}^{M} \frac{|B_m|}{n} \left| \text{acc}(B_m) - \text{conf}(B_m) \right|$,
where $n$ is the number of test triples, $B_m$ is the bin containing all predictions with confidence score in a given region of $[0, 1]$, 
acc($B_m$) measures the average link prediction accuracy in bin $B_m$,
and conf($B_m$) measures the average confidence score in bin $B_m$.
ECE is in [0, 1], and lower is better. 
For all reported ECE values, we use 10 bins. 

\begin{table}[t!]
\centering
\caption{Datasets used in our closed-world evaluation.}
\label{table:datasets}
\resizebox{0.9\columnwidth}{!}{
\begin{tabular}{r rrr}
    \toprule
     &  \# entities & \# relations & \# triples \\ 
     \toprule
     \wn{} & 40,493 & 11 & 93,003 \\ 
     \fb{} & 14,290 & 773 & 272,192 \\ 
     \bottomrule
\end{tabular}
\vspace{-.3cm}
}
\end{table}

\begin{table*}[!t]
\centering
	\caption{
		ECE (10 bins) and accuracy on \wn{} and \fb{}. 
		$\uparrow$: Higher is better.
        $\downarrow$: Lower is better. 
		% For ECE, lower is better. 
	}
	\label{table:relation-prediction}
	\resizebox{1\textwidth}{!}{
	\begin{tabular}{rr ccccccc c ccccccc }
	    \toprule
	     & &  \multicolumn{7}{c}{\wn{}} && \multicolumn{7}{c}{\fb{}} \\ 
	     \cline{3-9} \cline{11-17}
	     & & \multirow{2}{*}{Uncalib.} && \multicolumn{2}{c}{One-vs-all} && \multicolumn{2}{c}{Multiclass} &&  \multirow{2}{*}{Uncalib.} && \multicolumn{2}{c}{One-vs-all} && \multicolumn{2}{c}{Multiclass} \\ 
	     \cline{5-6} \cline{8-9} \cline{13-14} \cline{16-17} 
	     & &  && Platt & Iso. && Vector & Matrix && && Platt & Iso. && Vector & Matrix \\
	     \toprule
	     \multirow{4}{*}{ECE ($\downarrow$)} & \transe{}  & 0.624 && 0.054 & 0.040  && \textbf{0.014} & 0.022 && 0.795 && 0.071 & \textbf{0.016} && 0.026 & 0.084 \\ 
	     & \transh{}  &  0.054 && 0.057 & 0.044 && \textbf{0.018} &  0.027 &&  0.177 && 0.081 & \textbf{0.024} && 0.031 & 0.089 \\ 
	     & \distmult{}  & 0.046 && 0.040 & 0.029 && 0.044 &  \textbf{0.014} && 0.104 && 0.095 & 0.031 && \textbf{0.018} & 0.054 \\ 
	     & \complex{}  &  0.028 && 0.041 & 0.034 && 0.035 & \textbf{0.020}  && 0.055 && 0.102 & 0.037 && \textbf{0.024} & 0.112  \\ 
	     \midrule 
	     \multirow{4}{*}{Acc. ($\uparrow$)} & \transe{}  & 0.609 && 0.609 & 0.609 && 0.724 & \textbf{0.739} && 0.849 && 0.849 & 0.849 && \textbf{0.857} & 0.842 \\ 
	     & \transh{}  & 0.625 && 0.625 & 0.625 && 0.735 & \textbf{0.740} && 0.850 && 0.850 & 0.850 && \textbf{0.858} & 0.839 \\ 
	     & \distmult{}  & 0.570 && 0.570 & 0.570 && 0.723 & \textbf{0.761} && 0.819 && 0.819 & 0.819 && 0.862 & \textbf{0.871} \\ 
	     & \complex{}  & 0.571 && 0.571 & 0.571 && 0.750 & \textbf{0.781} && 0.884 && 0.884 & 0.884 && \textbf{0.908} & 0.892 \\ 
	     \bottomrule
	\end{tabular}
	}
\end{table*}

\begin{figure*}[t]
    \centering
    \begin{subfigure}[t]{0.99\columnwidth}
        \centering
        \includegraphics[width=0.9\textwidth]{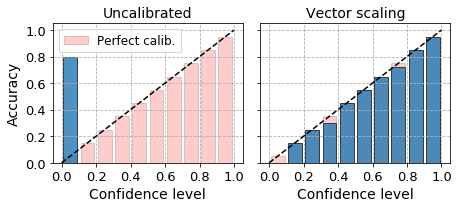}
        \caption{\transe{} }
        \label{fig:transe-reliability}
    \end{subfigure}
    ~
    \begin{subfigure}[t]{0.99\columnwidth}
        \centering
        \includegraphics[width=0.9\textwidth]{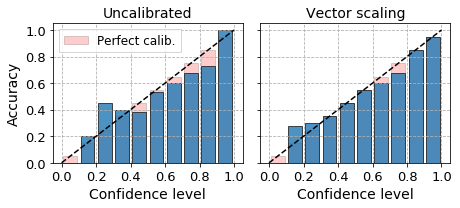}
        \caption{\distmult{} }
        \label{fig:distmult-reliability}
    \end{subfigure}%
    \caption{Reliability diagrams on \fb{} with predictions grouped into 10 bins.}
    \label{fig:reliability-closed}
\end{figure*}

\subsection{Data}
\label{sec:closed-data}

We use two link prediction benchmarks (Table~\ref{table:datasets}): The \textbf{\wn{}} semantic relation network~\cite{dettmers2018convolutional} and a version of the FB15K encyclopedic knowledge graph~\cite{bordes2013translating}.
We refer to this dataset as \textbf{\fb{}} because we link it to Wikidata~\cite{vrandevcic2014wikidata} to use as an external reference in \S~\ref{sec:open} for data annotation, discarding entities without entries in Wikidata.  
Following standard practice, we remove inverse relations from \fb{}, which artificially inflate link prediction accuracy~\cite{dettmers2018convolutional}.
We randomly split both datasets into 80/10/10 train/validation/test triples to ensure a sufficient number of validation triples for calibration. 

Note that there have been recent (concurrent) efforts to construct appropriate datasets for evaluating \kge{} calibration~\cite{pezeshkpour2020revisiting,safavi2020codex}.
Analysis on these new datasets is an important direction for future work.

\subsection{Results and discussion}
\label{sec:closed-discussion}

We implement our methods in an extension of the OpenKE library.\footnote{\url{https://github.com/thunlp/OpenKE/}}
To understand ``off-the-shelf'' calibration, we train models with the original loss functions and optimizers in the respective papers. 
Appendix~\ref{appx:implementation} provides details on implementation and model selection. 

\paragraph{Calibration error}
Table~\ref{table:relation-prediction} gives the ECE of all models before and after calibration using each technique in \S~\ref{sec:calibration-techniques}.
Confidence scores prior to calibration are scaled via the softmax. 
Across datasets, standard techniques \textbf{calibrate models within 1-2 percentage points of error} under the \cwa. 
In most cases, the strongest methods are the multiclass (softmax) approaches. 
The only exception is matrix scaling on \fb{}, which overfits due to the large number of classes in the dataset (773 in \fb{} versus only 11 in \wn, Table~\ref{table:datasets}). 
Evidently, taking the softmax over $k$ classes leads to more discriminative probabilities than setting up $k$ separate one-versus-all calibration problems and performing post-hoc normalization. 

We also observe that off-the-shelf calibration error is correlated with model type, as the bilinear models (\distmult{}, \complex{}) consistently have lower ECE than the translational models (\transe{}, \transh{}). 
To illustrate these differences, Figure~\ref{fig:reliability-closed} gives reliability diagrams for \transe{} and \distmult{} before and after calibration. 
Reliability diagrams~\cite{guo2017calibration} bin predictions by confidence level into equally-sized regions of [0, 1] and show the relationship between average confidence level and accuracy in each bin, similar to ECE (\S~\ref{sec:closed-task}). 
Without calibration, \transe{} is underconfident because it scores all predictions nearly the same, whereas \distmult{} is better calibrated.
We observe a similar pattern for \transh{} and \complex{}. 

One potential explanation for this difference is that multiplicative scoring functions lead to more discriminative scores due to the composition of dot products, which amplify embedding values. 
In fact, \transe{} is the only model that does not apply any dot product-based transformation to embeddings, leading to the worst off-the-shelf calibration. 
Another explanation relates to losses: 
All methods except \complex{} are trained with margin ranking loss, which optimizes the ordering of predictions rather than the values of prediction scores.
By contrast, \complex{} is trained with binary cross-entropy loss, the same loss that we use to calibrate models in the validation stage. 

\paragraph{Link prediction accuracy}
Table~\ref{table:relation-prediction} also compares link prediction accuracy before and after calibration. 
In most cases vector and matrix scaling improve accuracy, which is reminiscent of previous work showing that training \kge{} with softmax cross-entropy improves link prediction performance~\cite{kadlec2017knowledge,safavi2020codex}. 
We conclude that for relation prediction under the \cwa{}, \textbf{vector scaling provides the best trade-off} between calibration, accuracy, and efficiency, as it consistently improves accuracy and calibration with only $O(k)$ extra parameters.

\section{Open-world evaluation}
\label{sec:open}
We now address the more realistic \textbf{open-world assumption} (\owa), in which predictions not present in the knowledge graph are considered \emph{unknown}, rather than false, until ground-truth labels are obtained. 
While the \owa{} is beneficial because it helps us assess \kge{} calibration under more realistic conditions, it is also challenging because it significantly changes the requirements for evaluation. 
Specifically, now we need a label for every triple considered, whereas with the \cwa{} we only needed labels for a small group of positives. 

We emphasize that this is the reason the \owa{} is rarely used to evaluate \kge{}.
Narrowing down the large space of unknowns to a manageable smaller set and labeling these triples can be difficult and costly. 
We thus contribute first steps toward evaluation strategies under the \owa. 

\subsection{Task and metrics}
\label{sec:open-setup}

Similar to the link prediction task in \S~\ref{sec:closed-task}, we construct $(h, ?, t)$ queries from $(h, r, t)$ knowledge graph triples.
A \kge{} model then scores relations $r_i \in R$ to answer these queries. 
However, here we only consider completions $(h, r_i, t) \not\in G$, those for which the \emph{truth values are not known ahead of time}, which reflects how practitioners would use \kge{} to complete knowledge graphs in deployment settings. 
We use \fb{} as our dataset for this task because it is linked to Wikidata; we provide links to entities' Wikidata pages in our crowdsourced label collection process (\S~\ref{sec:open-data}). 

\begin{figure}[t!]
    \centering
    \includegraphics[width=1\columnwidth]
    {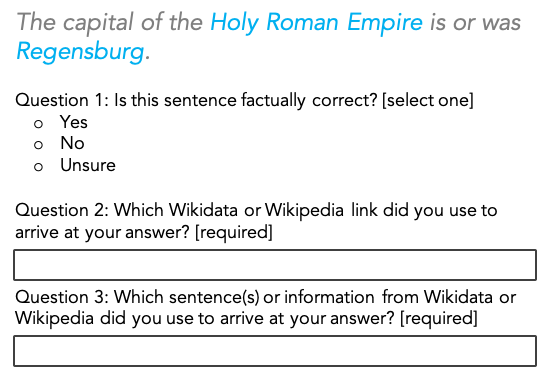}
    \caption{Open-world annotation interface.}
    \label{fig:open-example}
\end{figure}

\paragraph{Generating \owa{} predictions}
For each $(h, ?, t)$ query, we take the top-ranked $(h, \hat{r}, t)$ prediction made by a \kge{} model, and filter these predictions to unknowns $(h, \hat{r}, t) \not\in G$.

To simulate how a practitioner might narrow down a large set of unknowns to a few promising candidates under resource constraints (i.e., the cost of collecting labels), we take only the predictions made with confidence level $\geq 0.80$.
In other words, we choose to obtain \emph{many} judgments of a \emph{few} high-confidence predictions rather than \emph{few} judgments of \emph{many} lower-confidence predictions.
This helps us robustly compute agreement, maximize the probability of positives, and control quality.

We run this generation process for each \kge{} model from \S~\ref{sec:kge} trained on \fb{} before and after calibration.
We use vector scaling as our calibrator because it yields the best results on \fb{} under the \cwa{} (\S~\ref{sec:closed-discussion}). 

For evaluation, we use the same accuracy and calibration metrics as in \S~\ref{sec:closed-task}. 
However, since there is no ``test set'' in the open world, we must obtain ground-truth labels on predictions, discussed next. 

\begin{table*}[h!]
\centering
\renewcommand\thetable{5}
\caption{Examples of \owa{} predictions before and after calibration. 
}
\label{table:open-examples}
\resizebox{1\textwidth}{!}{
\begin{tabular}{ c lll rr c } 
\toprule
    & {Head $h$} & {Predicted relation $\hat{r}$} & {Tail $t$} & Model & Conf. $\hat{p}$ & True?  \\ 
    \toprule
    \multirow{5}{*}{\rotatebox[origin=c]{90}{Uncalib.}} & Bloomfield Hills &	/location/administrative\_division/second\_level\_division\_of & United States of America & \complex{}	& 0.985 & \textcolor{alizarin}{\xmark} \\ 
    & Spanish &	/language/human\_language/countries\_spoken\_in & Spain & \complex{} & 0.946 & \textcolor{cadmiumgreen}{\cmark} \\ 
    & New Hampshire &	/location/location/containedby & Hampshire & \distmult{}	& 0.860 & \textcolor{alizarin}{\xmark} \\ 
    & Billie Holiday & /music/artist/origin & New York City & \complex{} &	0.844 & \textcolor{cadmiumgreen}{\cmark} \\ 
    & egg & /food/ingredient/compatible\_with\_dietary\_restrictions & veganism & \complex{} & 0.844 & \textcolor{alizarin}{\xmark} \\ 
    \midrule
     \multirow{5}{*}{\rotatebox[origin=c]{90}{Vector}} & Asia & /locations/continents/countries\_within & Kazakhstan & \transh{} & 0.999 & \textcolor{cadmiumgreen}{\cmark} \\
    %  & House of Windsor & /royalty/monarch/royal\_line & Charles, Prince of Wales & \distmult{} & 0.999 & \textcolor{cadmiumgreen}{\cmark} \\
     & Shigeru Miyamoto	& /architecture/architectural\_style/architects & Mario \& Sonic at the Olympic Games & \distmult{} & 0.958 & \textcolor{alizarin}{\xmark} \\ 
     & Gujarati & /language/human\_language/countries\_spoken\_in & Uganda & \transe{} & 0.871 & \textcolor{cadmiumgreen}{\cmark} \\ 
     & Finnish & /location/location/containedby	& Europe & \transh{} & 0.843 & \textcolor{alizarin}{\xmark} \\
     & James Wong Jim & /people/person/nationality & Hong Kong & \complex{} & 0.832 & \textcolor{cadmiumgreen}{\cmark} \\
 \bottomrule
\end{tabular}
}
\end{table*}

\subsection{Data annotation}
\label{sec:open-data}

We collect judgments of the unknown $(h, \hat{r}, t) \not\in G$ predictions over \fb{} using the Figure 8 crowdsourcing platform.\footnote{\url{https://www.figure-eight.com/}} 
In the task, crowd workers answer whether each prediction is factually correct (Figure~\ref{fig:open-example}). 
Triples are presented as sentences, converted via pre-defined relation templates, with links to the Wikidata entries of the head and tail entities.
Appendix~\ref{appx:tf} gives sentence template examples, as well as more details on data preprocessing and the data annotation instructions. 

\paragraph{Participants}
We limit the annotation task to the highest-trusted group of contributors on Figure 8, and require references from Wikidata or Wikipedia for answers.
We also pre-label 20\pct{} of all triples and require participants to pass a 5-question ``quiz'' before starting the task and maintain 90\pct{} accuracy on the remaining gold questions.
We gather judgments for 1,152 triples, and collect five judgments per triple, taking the majority label as ground-truth. 
The inter-annotator agreement using Fleiss' kappa~\cite{fleiss1971measuring} is 0.7489 out of 1. 

\subsection{Results and discussion}
\label{sec:open-discussion}

\begin{table}[t!]
    \centering
    \renewcommand\thetable{4}
    \caption{ECE and link prediction accuracy by model in the open-world setting, before and after calibration. 
    The translational models do not make any predictions at a confidence level over 0.80 before calibration.
    }
    \label{table:open-acc-ece}
    \resizebox{0.9\columnwidth}{!}{
    \begin{tabular}{r cc c cc}
        \toprule
         & \multicolumn{2}{c}{ECE ($\downarrow$)} && \multicolumn{2}{c}{Accuracy ($\uparrow$)} \\ 
         \cline{2-3} \cline{5-6} 
         & Uncalib. & Vector && Uncalib. & Vector  \\ 
        \toprule
        \transe{}  & - & 0.234 && - & 0.594 \\
        \transh{}  & - & 0.307 && - & 0.521 \\
        \distmult{}  & 0.618 & 0.344 && 0.308 & 0.509 \\
        \complex{}  & 0.540 & 0.291 && 0.293 & 0.581 \\
        \midrule 
        Aggregate & 0.548 & 0.296 && 0.295 & 0.549 \\ 
        \bottomrule
    \end{tabular}
    }
\end{table}

Table~\ref{table:open-acc-ece} compares calibration error and link prediction accuracy before and after applying vector scaling. 
As shown in the table, the translational models do not make any uncalibrated predictions above a confidence level of 0.80 due to underconfidence, as dicussed in \S~\ref{sec:closed-discussion}.
The bilinear models, \distmult{} and \complex{}, are much less calibrated off-the-shelf than under the \cwa{} (c.f. Table~\ref{table:relation-prediction}). 

\begin{figure}[!t]
    \centering
    \includegraphics[width=0.75\columnwidth]{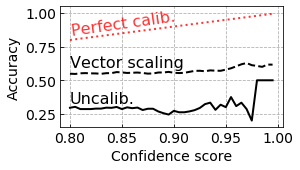}
    \caption{Reliability before and after calibration, aggregated across all four models. 
    }
    % \vspace{-.5cm}
    \label{fig:reliability-open}
\end{figure}

Even after vector scaling, which reduces ECE significantly for both models and scales the scores of the translational models appropriately, \textbf{all models are overconfident}, 
collectively reaching around 50-60\pct{} accuracy at the 80-100\pct{} confidence level (Table~\ref{table:open-acc-ece} and Figure~\ref{fig:reliability-open}). 
This is consistent with observations of \kge{} overconfidence made by~\citet{pezeshkpour2020revisiting} for the task of triple classification, as well as observations on the general overconfidence of neural networks for vision and language processing~\cite{guo2017calibration}. 

We also do not observe any correlation between a model's level of exposure to a particular relation type and its calibration on that relation type. 
For example, all models achieve relatively low ECE ($<4\pct{}$) on the relation \emph{\nolinkurl{ /language/human\_language/countries\_spoken\_in}}, which appears in only 0.148\pct{} of all triples in \fb{}. 
By contrast, for the relation \emph{\nolinkurl{/location/location/containedby}}, which appears in 2.30\pct{} of all \fb{} triples (15$\times$ more frequent), all models are poorly calibrated both before and after vector scaling (ECE $>10\pct{}$). 
We discuss these results and behaviors in more detail next.

\paragraph{Challenges of the \owa{}}
Accurately calibrating \kge{} models (and evaluating calibration thereof) is challenging under the \owa{} for several reasons.
First, in the \cwa{}, all queries are known to have at least \textbf{one correct answer ahead of time}, whereas in the \owa{} we have no such guarantee.
This highlights one of the fundamental challenges of the \owa{}, which is that of selecting predictions from a vast space of unknowns to maximize the probability of positives. 
It is likely that different strategies for selecting unknowns would lead to different observed levels of calibration. 

In the \owa{} there is also a \textbf{mismatch between negatives} in the calibration and evaluation stages. 
Recall that in the calibration stage, we take completions not seen in the graph as negative samples (\S~\ref{sec:calibration}), which is essentially a closed-world assumption.
By contrast, at evaluation time we make an open-world assumption. 
Higher-quality validation negatives may alleviate this problem; indeed, recent works have raised this issue and constructed new datasets toward this direction, albeit for the task of triple classification~\cite{pezeshkpour2020revisiting,safavi2020codex}. 

Finally, our observation about the varying levels of calibration per relation suggests that \textbf{some relations are simply more difficult to }calibrate because of the knowledge required to accurately model them. 
Most popular ``vanilla'' \kge{} models do not explicitly make use of external knowledge that can help refine prediction confidences, such as entity types, compositional rules, or text. 

Table~\ref{table:open-examples} provides examples of high-scoring  predictions made before and after calibration with corresponding labels. 
While most predictions are grammatically correct, it is perhaps not reasonable to expect \kge{} to capture certain types of semantics, logic, or commonsense using just the structure of the graph alone, for example that architects can design buildings but not video games (Table~\ref{table:open-examples}). 

\paragraph{Link prediction accuracy}
As shown in Table~\ref{table:open-acc-ece}, calibration with vector scaling on \fb{} {improves \owa{} link prediction accuracy by 20-28 percentage points}, which is significantly higher than under the \cwa{} (c.f. Table~\ref{table:relation-prediction}), in which it improved accuracy by 1-5 percentage points on \fb. 
We conclude that from a practitioner's perspective, vector scaling is a \textbf{practical technique for making predictions more accurate and trustworthy} even if it does not perfectly calibrate models.

\section{Case study}
\label{sec:fitb}
Finally, we conduct a case study of human-AI knowledge graph completion as a proof of concept on the benefits of \kge{} calibration for practitioners. 
In this experiment, given ``fill-in-the-blank'' sentences corresponding to incomplete knowledge graph triples, the task is to choose from multiple-choice answer lists generated by \kge{} to complete the sentences.
% Motivated by the idea that calibrated confidences can act as decision support, 
We show that, compared to annotators \emph{not} provided with confidence scores for this task, annotators provided with calibrated confidence scores for answer choices \textbf{more accurately and efficiently} complete triples.  

\subsection{Data}
\label{sec:fitb-data}

We construct a knowledge graph consisting of 23,887 entities, 13 relations, and 86,376 triples from Wikidata. 
We collect triples in which the head entity is categorized as a writer on Wikidata, and 13 people-centric relations (e.g., \emph{born in}, \emph{married to}).
We extract our dataset directly from Wikidata to guarantee that all answers are resolvable using a single public-domain source of information. 
We choose writing as a domain because it is less ``common knowledge'' than, e.g., pop culture. 

\paragraph{Task input}
After training and calibrating each \kge{} model from \S~\ref{sec:kge} over the Wikidata graph, we use our best-calibrated model (\complex{} + Platt scaling, ECE $<0.01$ under the \cwa{}) to predict relations. 
Per triple, we take the top-five predicted relations $\{\hat{r}\}_{i=1}^{5}$ and their calibrated confidence scores $\{\hat{p}\}_{i=1}^5$.
We filter these predictions to a sample of 678 triples by choosing only instances whose ground-truth relation $r$ is in the top-five predictions $\{\hat{r}\}_{i=1}^{5}$, balancing class proportions, and discarding questions with answers that are easy to guess.
Appendix~\ref{appx:fitb-data} provides more details.  

\begin{figure}
    \centering
    \includegraphics[width=1\columnwidth]
    {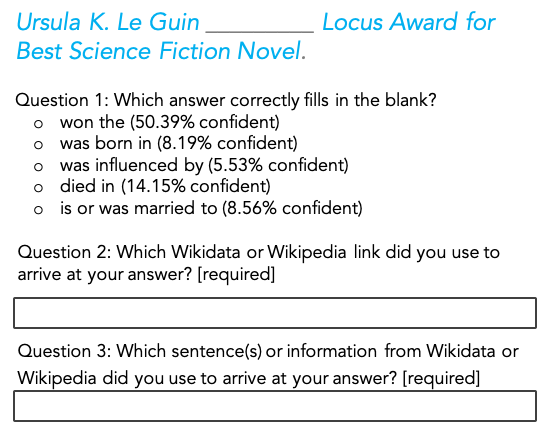}
    \caption{Example completion task from our case study.
    The confidence scores shown in parentheses for Question 1 are presented to the confidence group only. }
    \label{fig:fitb-example}
    % \vspace{-.5cm}
\end{figure}

\subsection{Task setup}
\label{sec:fitb-task}

The task is to complete triples by choosing the correct relation among the top-five \kge{}-predicted relations $\{\hat{r}\}_{i=1}^{5}$, presented in natural language. 

We conduct an A/B test whereby we vary how the confidence scores $\{\hat{p}\}_{i=1}^5$ for answer choices are presented to participants.
We provide the \textbf{no-confidence (control) group} with multiple-choice answers in their natural language form without any accompanying confidence estimates, whereas the \textbf{confidence (treatment) group} is provided a calibrated confidence score along with each answer candidate in parentheses (Figure~\ref{fig:fitb-example}). 
We also provide the confidence group with an extra paragraph of instructions explaining that the confidence scores are generated by a computer system; Appendix~\ref{appx:fitb-task-instructions} provides the full task instructions. 

To mitigate position bias, we randomize the presentation order of answer choices so that the answers are not necessarily ranked in order of confidence.
The answer candidates are presented in the \emph{same} randomized order for both groups. 

\paragraph{Participants}
We recruit 226 participants for the no-confidence group and 202 participants for the confidence group from Figure 8.
Participants are required to pass a 10-question ``quiz'' and maintain 50\pct{} minimum accuracy across all pre-labeled gold questions. 
We limit each participant to up to 20 judgments, and  collect three judgments per triple.

% Post-task, we present an optional satisfaction survey on: 
% \textbf{(1)}~The ease of the task; 
% \textbf{(2)}~The fairness of the pre-labeled gold questions; \textbf{(3)}~The level of pay relative to other tasks; \textbf{(4)}~The clarity of the instructions; and \textbf{(5)}~Overall satisfaction. 
% All answers are given on a scale of one to five points. 

\begin{table}[t]
\centering
\caption{
Case study results.
$\uparrow$: Higher is better.
$\downarrow$: Lower is better. 
\textbf{Bold}: Significant at $p <0.05$.
\underline{Underline}: Significant at $p<0.01$.
$^*$: $p$-value not applicable. 
Detailed explanations are given in \S~\ref{sec:fitb-discussion}. 
}
\label{table:fitb-results}
\resizebox{1\columnwidth}{!}{
\def\arraystretch{1}
\begin{tabular}{ r ccc cc } 
\toprule
    & \multicolumn{3}{c}{Accuracy $\uparrow$} && Sec. per  \\
    \cline{2-4} 
    & Overall & Per triple  & Per person && triple $\downarrow$  \\ 
    \toprule
    No-conf. & 0.8977 & 0.8969 & 0.9120 & & 36.88 \\ 
    Conf. & {0.9175}$^*$ & \underline{\textbf{0.9220}} & \underline{\textbf{0.9478}} & & \textbf{31.91} \\ 
    \midrule
    Abs. diff. & +0.0198 & +0.0251 & +0.0358 & & -4.97 \\
    Rel. diff. & +2.21\pct{} & +2.79\pct{}  & +3.93\pct{} & & -13.48\pct{} \\   
 \bottomrule
\end{tabular}
}
\end{table}

\subsection{Results and discussion}
\label{sec:fitb-discussion}

Table~\ref{table:fitb-results} summarizes the results of our case study. 
For the accuracy results, statistical significance is determined with the Wilcoxon rank-sum test~\cite{wilcoxon1992individual} due to non-normality. 
For the efficiency results, statistical significance is determined with an independent two-sample $t$-test. 

\paragraph{Accuracy}
The proportion of correct judgments in the no-confidence group was 0.8977 compared to 0.9175 in the confidence group, an improvement of 1.98 percentage points. 
In terms of the average judgment accuracy per triple,
or the number of correct judgments divided by the number of judgments per triple, 
the no-confidence and confidence averages were 0.8969 and 0.9220 respectively, a significant difference ($p<10^{-3}$). 
The average judgment accuracy per participant also differed significantly ($p < 10^{-6}$), again in favor of the confidence group. 

Finally, model accuracy was 0.6268, meaning that for 62.68\pct{} (425/678) of triples seen by participants in the confidence group, the answer choice with the highest confidence score was the correct answer. 
Given that the confidence group's accuracy was much higher (0.9175 versus 0.6268), we can conclude that the participants in this group did not blindly trust the confidence scores.

\paragraph{Efficiency}
For this comparison we remove outliers with average judgment times more than two standard deviations away from the group mean. %  to account for participants who paused working in the middle of the task. 
The mean time per judgment was 36.88 seconds in the no-confidence group (194 participants) versus 31.91 seconds in the confidence group (179 participants), a significant difference ($p=0.010$).
Note that we required sources and textual references for all answers across both groups (Questions 2 and 3 in the example in Figure~\ref{fig:fitb-example}).
However, even with these quality control measures, the confidence group was significantly faster.
% used quality control measures (Questions 2 and 3, Figure~\ref{fig:fitb-example}) 
% may have mitigated some of the differences in efficiency between the two groups.

% \paragraph{Satisfaction}
% We obtained 15 responses to the satisfaction survey (\S~\ref{sec:fitb-task}) in the control group and 32 responses in the treatment group (note that contributor satisfaction surveys are strictly optional on Figure 8). 
% Those assigned to the treatment group responded to all questions with higher average scores (Table~\ref{table:fitb-results}).
% The difference in ``overall satisfaction'' was the most significant ($p=2.9\times 10^{-3}$). 

In conclusion, the results of our case study indicate that human-AI knowledge graph completion is more accurate and efficient with calibrated confidence scores generated by \kge{}. 
These findings suggest that calibrated probabilities are indeed trustworthy to practitioners, motivating the utility of calibration for human-AI tasks. 

\section{Conclusion}
\label{sec:conclusion}
We investigate calibration as a technique for improving the trustworthiness of link prediction with \kge{}, and uniquely contribute both closed-world \emph{and} open-world evaluations; the latter is rarely studied for \kge{}, even though it is more faithful to how practitioners would use \kge{} for completion. 
We show that there is significant room for improvement in calibrating \kge{} under the \owa{}, and motivate the importance of this direction with our case study of human-AI knowledge graph completion. 
As knowledge graphs are increasingly used as gold standard data sources in artificial intelligence systems, our work is a first step toward making \kge{} predictions more trustworthy. 

\section*{Acknowledgments}
We thank Caleb Belth for his feedback. 
This material is partially supported by the National Science Foundation under Grant No. IIS 1845491,  Army Young Investigator Award No. W911NF1810397, and an NSF Graduate Research Fellowship. 

\balance
\bibliography{references}
\bibliographystyle{acl_natbib}

\clearpage
\appendix
\begin{table*}[t!]
\centering
\caption{Example sentence templates for relations in \fb{}.
$h$: Head entity label.
$t$: Tail entity label. 
}
\label{table:fb-relation-templates}
\resizebox{\textwidth}{!}{
\begin{tabular}{ l l l } 
\toprule
{Relation} & {Template} & Reverse template \\ 
\toprule
    /architecture/structure/architect & $h$ was designed by the architect $t$. & $t$ was the architect of $h$. \\ 
    % /aviation/airline/hubs & $h$ is an airline with a hub at $t$. & $t$ is a hub for the airline $h$.  \\
     /cvg/computer\_videogame/sequel & $h$ is a video game with sequel $t$. & $t$ is the sequel of the video game $h$. \\
    %  /education/fraternity\_sorority/colleges\_and\_universities & $h$ is a fraternity or sorority at $t$. & $t$ is a school with the fraternity or sorority $h$. \\ 
     /fight/crime\_type/victims\_of\_this\_crime\_type & $h$ is a crime that happened to $t$. & $t$ was a victim of $h$. \\ 
     /film/writer/film & $h$ was a writer on the film $t$. & The film $t$'s writers included $h$. \\
    %  /finance/currency/countries\_formerly\_used & $h$ is the former currency of $t$. & $t$ formerly used the $h$ currently. \\
     /food/diet\_follower/follows\_diet & $h$ follows a diet of $t$. & $t$ is a diet followed by $h$. \\
    %  /geography/river/mouth & $h$ is a river with a mouth into the $t$.  \\ 
     /government/government\_agency/jurisdiction & $h$ is a governmental agency with jurisdiction over $t$. & $t$ is under the jursidiction of $h$.  \\ 
    %  /internet/website/category & $h$ has a website in the category of $t$. \\ 
    %  /language/human\_language/main\_country & The main country in which $h$ is spoken is $t$. \\ 
    /medicine/risk\_factor/diseases & $h$ has the risk of causing $t$. & $t$ can be caused by $h$. \\
    % /music/artist/origin & $h$ is or was a musician from $t$. &  \\ 
     /people/person/nationality & $h$ has or had $t$ nationality. & $t$ is the nationality of $h$. \\ 
    %  /religion/religion/branched\_from & $h$ is a religion that branched from $t$. & $t$ is a religion from which $h$ branched. \\ 
    %  /sports/sports\_league/sport & $h$ is a $t$ sports league. & The sport of $t$ \\ 
     /time/holiday/featured\_in\_religions & $h$ is a holiday featured in the religion of $t$. & $t$ is a religion that celebrates $h$. \\ 
    %  /tv/tv\_program/country\_of\_origin & $h$ is a TV program originating in $t$. \\ 
 \bottomrule
\end{tabular}
}
\end{table*}

\section{Implementation details}
\label{appx:implementation}

To select models, we grid search over the number of training epochs in \{200, 300, 500\},
the batch size in \{100, 200, 500\}, and the embedding dimension in \{50, 100\}. 
For training, we use random uniform negative sampling to speed up the training process.
% and leave more sophisticated negative sampling approaches (e.g.,~\cite{cai2018kbgan}) to future work.
We search over the number of negative relations sampled per positive triple in \{1, 5\}. 

We follow the original papers' choices of loss functions and optimizers. 
For loss functions, we use margin ranking for \transe, \transh, and \distmult{} and binary cross-entropy for \complex{}, and grid search over the margin hyperparameter in \{1, 5, 10\} for margin ranking. 
For optimizers, we use SGD for \transe{} and \transh{}, and Adagrad for \distmult{} and \complex{}, with a learning rate of 0.01.

We use the scikit-learn implementations of one-versus-all Platt scaling and isotonic regression\footnote{\url{https://scikit-learn.org/stable/modules/generated/sklearn.calibration.CalibratedClassifierCV.html}}, and implement vector and matrix scaling in Tensorflow with L-BFGS~\cite{liu1989limited} limited to 2,000 iterations following the reference implementation provided by~\citet{guo2017calibration}.\footnote{\url{https://github.com/gpleiss/temperature\_scaling}}
% The one-versus-all approaches run within a few minutes, whereas matrix scaling takes up to a few hours due to the larger number of parameters, on a single CPU.

\section{Open-world evaluation}
\label{appx:tf}

\subsection{Data}
\label{appx:tf-data}

To construct the set of triples for annotation, 
% in our \owa{} evaluation, we discard % lengthy compound Freebase relations that are difficult to translate into simple English phrases, e.g., \emph{\nolinkurl{/travel/transport\_terminus/travel\_destinations\_served./travel/transportation/transport\_terminus}}. 
% We also remove 
we discard relations pertaining to Netflix (e.g., \emph{/media\_common/netflix\_genre/titles}) to avoid disagreement due to crowd workers' countries of origin, since Netflix title availability varies widely by country.
We convert all triples to sentences with a set of pre-defined relation templates.
Because all relations can be reversed---e.g., (\emph{Beyonc\'{e}}, \emph{citizenOf}, \emph{USA}) and (\emph{USA}, \emph{hasCitizen}, \emph{Beyonc\'{e}}) express the same fact---we create two sentence templates for each relation and take the sentence that expresses the more plausible and/or grammatical statement per triple. 
Table~\ref{table:fb-relation-templates} gives examples of sentence templates for relations in \fb{}.

\subsection{Task instructions}
\label{appx:tf-instructions}

This section gives the data annotation task instructions. 
Note that we conduct two separate annotation tasks: One with links to entities' Wikidata pages, and one with links to entities' IMDb pages for \emph{/film} relations only (Wikidata is linked to both Freebase and IMDb). 
The instructions are exactly the same between the two versions of the task, except that each instance of ``Wikidata and/or Wikipedia'' is replaced with ``IMDb'' in the latter. 

\paragraph{Overview}
The goal of this task is to determine whether a given sentence is true or false.

\paragraph{Instructions}
Given a sentence that states a potentially true fact about the world, for example
\begin{quote}
    \emph{Elizabeth Alexandra Mary Windsor is the queen of the Commonwealth.}
\end{quote} 
Read the sentence carefully and answer whether the sentence is factually correct by choosing one of \emph{Yes}, \emph{No}, or \emph{Unsure}.
To arrive at your answer, you \textbf{must use English-language Wikidata and/or Wikipedia}, even if you know the answer ahead of time. Each sentence already contains links to potentially relevant Wikidata pages; however, if you do not find an answer in the Wikidata page, you must check related Wikipedia pages. You \textbf{may not use any external data sources beyond English-language Wikidata or Wikipedia}.
After you select your answer (Question 1), give the primary English-language Wikidata or Wikipedia URL (Question 2) and the text snippet or reasoning you used to arrive at your answer (Question 3).

\paragraph{Rules and Tips}

\begin{itemize}
    \setlength\itemsep{0.0001em}
    \item {Read each sentence carefully and check both Wikidata and Wikipedia before choosing your answer.}
    \item \textbf{Question 1}: If a sentence is not grammatically correct, treat it as false. If a sentence is grammatically correct but you cannot find any information on Wikidata or Wikipedia supporting or disproving its claim, or you cannot reason about whether its claim is true or false, choose Unsure.
    \item \textbf{Question 2}: You must copy-paste the primary Wikidata or Wikipedia link that you used to arrive at your answer. \textbf{Only copy-paste the single link that contains the most complete answer to the question.} You may use the provided Wikidata links, but you \textbf{may also need to check related Wikipedia pages if you do not find what you are looking for. You may not use any external data sources beyond English-language Wikidata or Wikipedia.}
    \item \textbf{Question 3}: You may copy-paste relevant textual snippets from Wikidata or Wikipedia. If there is no relevant text to copy-paste, you may write a brief explanation of how you arrived at your answer.
\end{itemize}

\paragraph{Examples}
We give two examples presented to crowd workers in the task instructions.

\begin{enumerate}
    \setlength\itemsep{0.0001em}
    \item \emph{\href{https://www.wikidata.org/wiki/Q134068}{Nawaz Sharif} is or was a leader of \href{https://www.wikidata.org/wiki/Q843}{Pakistan}.}
    \begin{itemize}
        \item Is this sentence factually correct?
        \begin{itemize}
            \item Yes
        \end{itemize}
        \item Which Wikidata or Wikipedia link did you use to arrive at your answer?
        \begin{itemize}
            \item \url{https://en.wikipedia.org/wiki/Nawaz\_Sharif}
        \end{itemize}
        \item Which sentence(s) or information from Wikidata or Wikipedia did you use to arrive at your answer?
        \begin{itemize}
            \item ``Mian Muhammad Nawaz Sharif is a Pakistani businessman and politician who served as the prime minister of Pakistan for three non-consecutive terms'' - from the Wikipedia page of Nawaz Sharif
        \end{itemize}
    \end{itemize}
    \item \emph{The capital of \href{https://www.wikidata.org/wiki/Q142}{France} is or was \href{https://www.wikidata.org/wiki/Q6397}{Avignon}.}
    \begin{itemize}
        \item Is this sentence factually correct?
        \begin{itemize}
            \item No
        \end{itemize}
        \item Which Wikidata or Wikipedia link did you use to arrive at your answer?
        \begin{itemize}
            \item \url{https://en.wikipedia.org/wiki/List\_of\_capitals\_of\_France}
        \end{itemize}
        \item Which sentence(s) or information from Wikidata or Wikipedia did you use to arrive at your answer?
        \begin{itemize}
            \item Avignon is not listed as a capital of France on the Wikipedia page about the capitals of France.
        \end{itemize}
    \end{itemize}
\end{enumerate}

\section{Case study}
\label{appx:fitb}

\subsection{Data}
\label{appx:fitb-data}

To convert all triples into natural language for the task, we map each relation in the dataset to a phrase: 
P19 (\emph{was born in}), P20 (\emph{died in}), 
P21 (\emph{is of gender}), P26 (\emph{is or was married to}), P101 (\emph{works or worked in the field of}), P103 (\emph{speaks or spoke the native language of}), P106 (\emph{works or worked as a}), P119 (\emph{is buried in}), P136 (\emph{created works in the genre of}), P140 (\emph{follows or followed the religion}), P166 (\emph{was awarded the}), P551 (\emph{lives or lived in}), and P737 (\emph{was influenced by}).

We train each model on all triples that we collected from Wikidata, but limit the task input to a subset of triples for which the correct answer is unambiguous but also not easy to guess.
To this end, we discard triples with relations that can be guessed via type matching: \emph{Gender} (the tail entity is always \emph{male} or \emph{female} in our dataset), \emph{award received} (the tail entity usually contains the word ``award'', ``prize'', etc.), and \emph{place of burial} (the tail entity usually contains the word ``cemetery'').
We also discard triples with relations that can be interpreted as synonyms of one another (\emph{occupation} and \emph{genre}, e.g., ``fiction writer''), and triples with the relation \emph{field of work} for which the tail entity is synonymous with ``writer'' or ``author'', since all people in the dataset are categorized as authors or writers on Wikidata. 
Finally, we remove triples for which there is more than one correct answer in the top-five predicted relations. 

\subsection{Task instructions}
\label{appx:fitb-task-instructions}

This section gives the task instructions of the case study. 
\ul{Underline} indicates that the enclosed text was presented to the confidence group only. 

\paragraph{Overview}
The goal of this task is to complete a sentence so that it states a true fact about the world.

\paragraph{Instructions}
Given a partially complete sentence, fill in the blank with exactly one of the provided answer choices so that the sentence states a true fact about the world. To arrive at your answer, you \textbf{must use the provided Wikidata links in each sentence. You may not use any external data sources beyond the provided Wikidata links in each sentence.}
\ul{Please note that we have used a computer system to generate ``confidence values'' for each answer choice in order to help you with the task. These values signify our system's belief about which answer is most likely to be correct.}
After you select your answer (Question 1), give the single Wikidata URL (Question 2) and the text snippet or reasoning you used to arrive at your answer (Question 3). You must provide all answers in English.

\paragraph{Rules and Tips}
\begin{itemize}
    \item \textbf{Question 1}: Choose the answer that makes the sentence grammatically correct and factual according to Wikidata. Every sentence has at least one correct answer. If you believe a sentence has multiple equally correct answers, choose any of them.
    \item \textbf{Question 2}: You must copy-paste the single, entire Wikidata link that you used to arrive at your answer. The link that you copy-paste \textbf{must contain the correct answer that fills in the blank in the sentence}. You must use the Wikidata links provided in each sentence. You may not use any external data sources beyond the provided Wikidata links.
    \item \textbf{Question 3}: You may copy-paste relevant textual snippets from Wikidata. If there is no relevant text to copy-paste, you must write a brief explanation of how you arrived at your answer. You must provide all answers in English.
\end{itemize}

\paragraph{Examples} 
We give two examples presented to crowd workers in the task instructions.

\begin{enumerate}
    \item \emph{\href{https://www.wikidata.org/wiki/Q80440}{Anna Akhmatova} \underline{\hspace{3cm}} \href{https://www.wikidata.org/wiki/Q7243}{Leo Tolstoy}.}
        \begin{enumerate}
            \item was or is married to \ul{(40\% confident)}
            \item \textbf{was influenced by} \ul{(45\% confident)}
            \item was the academic advisor of \ul{(5\% confident)}
            \item was the child of \ul{(5\% confident)}
            \item was the parent of \ul{(5\% confident)}
        \end{enumerate}
        \begin{itemize} 
            \item Which Wikidata link did you use to arrive at your answer?
                \begin{itemize}
                    \item \url{https://www.wikidata.org/wiki/Q80440}
                \end{itemize}
            \item Which sentence(s) or information from Wikidata did you use to arrive at your answer?
                \begin{itemize}
                    \item Wikidata says that Anna Akhmatova was influenced by Leo Tolstoy.
                \end{itemize}
        \end{itemize}
    \item \emph{\href{https://www.wikidata.org/wiki/Q181659}{Ursula K. Le Guin} \underline{\hspace{3cm}} \href{https://www.wikidata.org/wiki/Q1056251}{Hugo Award for Best Short Story}.}
    \begin{enumerate}
        \item \textbf{was awarded the \ul{(40\% confident})}
        \item was influenced by \ul{(0\% confident)}
        \item created the \ul{(50\% confident)}
        \item was or is married to \ul{(10\% confident)}
        \item lives in \ul{(0\% confident)}
    \end{enumerate}
    \begin{itemize}
        \item Which Wikidata link did you use to arrive at your answer?
        \begin{itemize}
            \item \url{https://www.wikidata.org/wiki/Q181659}
        \end{itemize}
        \item Which sentence(s) or information from Wikidata did you use to arrive at your answer?
        \begin{itemize}
            \item Wikidata says that Ursula K Le Guin won the Hugo Award for Best Short Story.
        \end{itemize}
    \end{itemize}
\end{enumerate}

% \begin{table*}[t]
% \centering
% \caption{
% Case study results.
% $\uparrow$: Higher is better.
% $\downarrow$: Lower is better. 
% \textbf{Bold}: Significant at $p <0.05$.
% \underline{Underline}: Significant at $p<0.01$.
% $^*$: $p$-value not applicable. 
% Detailed explanations are given in \S~\ref{sec:fitb-discussion}. 
% }
% \label{table:fitb-results}
% \resizebox{1\textwidth}{!}{
% \def\arraystretch{1}
% \begin{tabular}{ r ccc ccc ccccc } 
% \toprule
%     & \multicolumn{3}{c}{\textbf{Accuracy $\uparrow$}} && \multirow{2}{*}{\textbf{Sec/judgment $\downarrow$}} && \multicolumn{5}{c}{\textbf{Satisfaction (points out of 5) $\uparrow$}}  \\
%     \cline{2-4}  \cline{8-12}
%     & Overall & Per triple  & Per worker &&  && Ease & Fairness & Pay & Clarity & Overall \\
%     \toprule
%     \textbf{No-confidence} & 0.8977 & 0.8969 & 0.9120 & & 36.88 & & 3.200 & 3.333 & 3.333 & 3.733 & 3.200 \\ 
%     \textbf{Confidence} & {0.9175}$^*$ & \underline{\textbf{0.9220}} & \underline{\textbf{0.9478}} & & \textbf{31.91} & & {3.500} & \textbf{3.742} & \textbf{3.875} & {4.031} & \underline{\textbf{4.000}} \\ 
%     \midrule
%     \textbf{Absolute diff.} & +0.0198 & +0.0251 & +0.0358 & & -4.97 & & +0.300 & +0.409 & +0.542 & +0.298 &  +0.800 \\
%     \textbf{Relative diff.} & +2.21\pct{} & +2.79\pct{}  & +3.93\pct{} & & -13.48\pct{} & & +9.37\pct{} & +12.26\pct{} & +12.50\pct{} & +6.29\pct{} & +25.00\pct{} \\   
%  \bottomrule
% \end{tabular}
% }
% \end{table*}

\end{document}